%% file: root.tex
\documentclass[letterpaper, 10 pt, conference]{ieeeconf}  

\input{sec/0-defs}
\renewcommand{\baselinestretch}{0.98}

\title{\textbf{Dexterous Manipulation Policies from RGB Human Videos} \\\textbf{via 3D Hand-Object Trajectory Reconstruction}
}
\author{Hongyi Chen$^{1}$, Tony Dong$^{1}$, Tiancheng Wu$^{1}$, Liquan Wang$^{2}$, Yash Jangir$^{1}$, Yaru Niu$^{1}$, \\ Yufei Ye$^{3}$, Homanga Bharadhwaj$^{1}$, Zackory Erickson$^{1,\dagger}$, Jeffrey Ichnowski$^{1,\dagger}$\\ \\
$^{1}$ Carnegie Mellon University,
$^{2}$ Georgia Institute of Technology,
$^{3}$ Stanford University, $^{\dagger}$ Equal advising \\ \\
\href{https://videomanip.github.io}{\texttt{videomanip.github.io}}\\}
\begin{document}

\makeatletter
\let\@oldmaketitle\@maketitle
\renewcommand{\@maketitle}{\@oldmaketitle
\centering
  \includegraphics[width=0.98\textwidth]{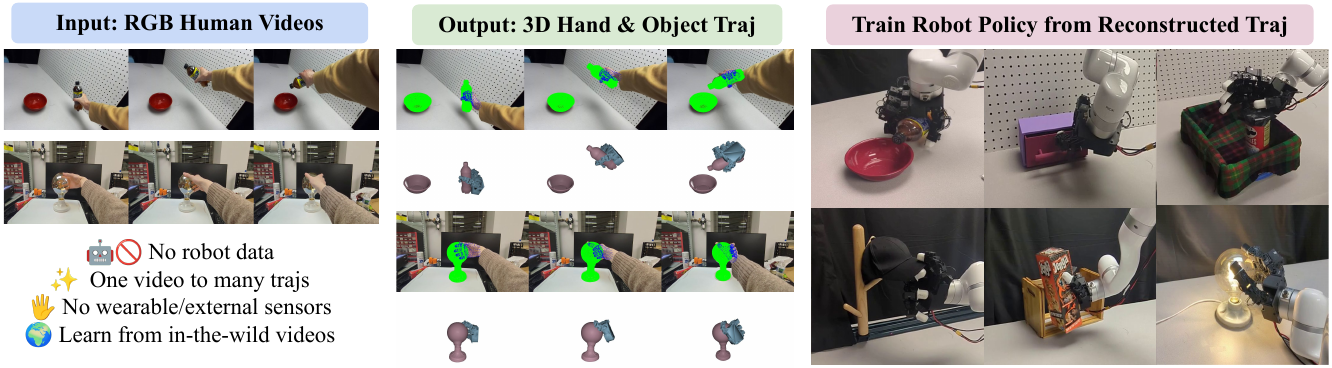}
  \captionof{figure}{\small Given RGB human videos, including both in-scene and in-the-wild recordings (left), the proposed \mysystem{} framework reconstructs trajectories of hand and object meshes (middle, projected onto RGB frames) and retargets them into robot actions to train policies. This enables dexterous manipulation learning directly from human videos without robot data or wearable/external sensing devices (right), offering a scalable alternative to conventional multifinger robot data collection pipelines.} 
  \label{fig:teaser}
  \vspace{-0.10in}
}
\makeatother

\let\oldmaketitle\maketitle
\renewcommand{\maketitle}{
  \oldmaketitle
  \addtocounter{figure}{-1}  
}

\maketitle

\input{sec/1-abstract}
\input{sec/2-intro}
\input{sec/3-related}
\input{sec/method}

\input{sec/experiment}

\section{Conclusion, Limitations, and Future Work}
\label{sec:conclusion}
We present a reconstruction-based framework that recovers 3D hand–object trajectories from RGB-only, in-the-wild human videos at low cost, without relying on specialized sensing devices. The reconstructed trajectories are shown to be effective for training dexterous grasping and manipulation policies, and provide more physically grounded supervision compared to retargeting-based approaches that rely on generated videos.

Despite these advantages, \mysystem{} has several limitations. 
(1) \mysystem{} framework relies on multiple 3D vision models, including image-to-mesh reconstruction and object pose estimation, which may introduce compounding errors across stages. We currently circumvent this issue by recording the human videos from approximately egocentric viewpoints that facilitate reliable hand and object reconstruction, instead of downloading generic videos from the internet. In future work, we plan to incorporate a trajectory verification module to automatically detect and filter erroneous reconstructions, and scale to more in-the-wild videos. 
(2) The current framework assumes static or approximately static camera setups. We plan to extend our approach to dynamic camera settings by leveraging recent advances in dynamic scene reconstruction~\cite{wang2025c4d, yu2025dyn}. 
(3) The manipulation policy learning currently relies on synthesized trajectories from DemoGen represented as 3D point clouds. In real-world manipulation, however, object point clouds are often difficult to track due to occlusions caused by the robot hand. We currently assume that the hand–object relative contact established during grasping is preserved throughout execution to update the object point clouds along with robot hand point clouds. In future work, we plan to explore alternative trajectory synthesis modalities beyond point clouds, such as image-based representations.




\bibliographystyle{IEEEtran}
\bibliography{IEEEabrv, references}

\end{document}

%% file: sec/0-defs.tex
\IEEEoverridecommandlockouts                              

\usepackage{graphics} 
\usepackage{epsfig} 
\usepackage{times} 
\usepackage{amsmath} 
\usepackage{amssymb}  
\usepackage{amsmath,amsfonts} 
\usepackage{url}
\usepackage{comment}
\usepackage{makecell, multirow}
\usepackage{booktabs}
\usepackage{pifont}
\usepackage{amsmath}
\usepackage{array}
\usepackage{xcolor}
\definecolor{linkbrown}{RGB}{120,70,20} 

\usepackage[pagebackref,breaklinks,colorlinks,allcolors=linkbrown]{hyperref}
\usepackage{mathrsfs}
\usepackage{amssymb}
\usepackage{amsfonts}
\usepackage{amsmath}
\usepackage{cleveref}
\usepackage[flushleft]{threeparttable}
\usepackage{tablefootnote}
\usepackage{hhline}
\usepackage{graphicx}
\usepackage{graphicx, subcaption}
\usepackage{hanging}
\usepackage{color}
\usepackage{tikz}
\usepackage{float,wrapfig}
\usetikzlibrary{calc,arrows,decorations.markings}
\usepackage{balance}
\usepackage{hhline}
\usepackage{algorithm}
\usepackage{mathtools}
\usepackage{algpseudocode}
\usepackage{svg}
\usepackage{ifthen}
\let\labelindent\relax
\usepackage[inline]{enumitem}
\usepackage{xspace}
\usepackage{amssymb}
\usepackage{tikz}
\usetikzlibrary{positioning, fit, backgrounds, calc, shapes.geometric, arrows.meta}
\usepackage{pifont} 
\usepackage{xcolor}
\usepackage{amssymb}
\usepackage{array} 

\definecolor{mygreen}{RGB}{0, 140, 0}       
\definecolor{highlightbg}{RGB}{235, 255, 235} 
\definecolor{gridgray}{RGB}{200, 200, 200}  
\definecolor{textgray}{RGB}{80, 80, 80}     
\definecolor{axiscolor}{RGB}{0, 70, 70}

\newcommand{\mysystem}{\textsc{VideoManip}\xspace}

%% file: sec/1-abstract.tex
\begin{abstract}
Multi-finger robotic hand manipulation and grasping are challenging due to the high-dimensional action space and the difficulty of acquiring large-scale training data.
Existing approaches largely rely on human teleoperation with wearable devices or specialized sensing equipment to collect robot demonstrations, limiting scalability. 
In this work, we propose \mysystem, a device-free framework that learns dexterous multi-finger manipulation directly from RGB human videos. Leveraging recent advances in computer vision, \mysystem reconstructs explicit robot-object trajectories from monocular videos by estimating human hand poses, object meshes, and retargets the reconstructed human motions to robot hands for manipulation learning. 
To make the reconstructed robot data suitable for dexterous manipulation training, we introduce hand-object contact optimization with interaction-centric grasp modeling, as well as a demonstration synthesis strategy that generates diverse training trajectories from a single video, enabling generalizable policy learning without additional robot demonstrations. 
In simulation, the learned grasping model achieves a 70.25\% success rate across 20 diverse objects using the Inspire Hand. In the real world, manipulation policies trained from RGB videos achieve an average 62.86\% success rate across seven tasks using the LEAP Hand, outperforming retargeting-based methods by 15.87\%.
Project videos are available at \href{https://videomanip.github.io}{videomanip.github.io}.
\end{abstract}

%% file: sec/2-intro.tex
\section{Introduction}
\label{sec:introduction}	
\emph{To what extent can dexterous robot manipulation be learned using human video-only supervision?}
Existing works have trained manipulation policies through human data obtained via hand-held grippers~\cite{liu2025fastumi}, wearables such as smart glasses~\cite{guzey2025dexterity} and headsets~\cite{jiang2025dexmimicgen, niu2025human2locoman}, or multi-camera studio setups~\cite{fu2025gigahands}. Although effective, these methods typically require specialized hardware, controlled capture environments, or direct human involvement, limiting scalability.
In contrast, RGB videos are ubiquitous and available at internet scale, offering the potential for scalable learning without specialized hardware. However, extracting accurate supervision for robot-object interactions from such data remains challenging, as these videos lack robot actions and precise 3D information.
In this work, we propose a framework for learning dexterous manipulation from RGB-only human videos alone. We evaluate this framework under progressively relaxed data conditions, from in-scene videos to in-the-wild videos.

Prior work explored leveraging egocentric human videos for manipulation learning; however, these approaches either have not been successfully deployed on multi-fingered robotic hands~\cite{bao2024handsonvlm, bharadhwaj2024track2act}, rely on robot demonstrations to fine-tune models pre-trained on in-the-wild dataset~\cite{shaw_internetvideos, singh2024videos, tao2025dexwild}, are trained on videos collected in deployment-specific scenarios~\cite{lum2025crossing}, or assume access to pre-scanned object models~\cite{qin2022dexmv}. 
Several studies reconstruct human hand poses or object meshes from images or videos of humans manipulating objects. However, these methods either have not demonstrated effective use of the reconstructed information for learning dexterous robotic hand manipulation~\cite{wu2024reconstructing, chen2025web2grasp, yu2025real2render2real}, or fail to exploit the reconstructed trajectories for direct imitation learning, instead relying on designed reward functions for manipulation policy learning~\cite{ye2025video2policy} or requiring additional policy fine-tuning~\cite{singh2024videos}.
In parallel, another line of research focuses on human teleoperation of robotic hands~\cite{chi2024universal, zhao2023learning}, often requiring head-mounted vision devices or specialized sensing equipment to track human hands~\cite{niu2025human2locoman, jiang2025dexmimicgen, qiu2025humanoid}. These dependencies on specialized hardware limit scalability and restrict data diversity.

To address these challenges, we propose \mysystem, a framework that learns dexterous grasping and manipulation entirely from RGB human videos, without requiring wearables or external sensor devices, robot demonstrations for post-training, or pre-scanned object meshes (Fig.~\ref{fig:teaser}). Leveraging recent advances in computer vision, such as metric depth estimation and image-to-mesh reconstruction, \mysystem{} recovers explicit 3D hand-object trajectories from monocular videos by estimating human hand poses, object meshes, and object scales, and subsequently retargets the reconstructed human motions to robot hands. For \emph{in-scene videos}, we assume known camera-robot extrinsic calibration and use it to transform reconstructed trajectories from the camera frame to the robot base frame for policy training. In contrast, for \emph{in-the-wild videos}, which are recorded outside robot setups without extrinsic calibration, we estimate the gravity direction from visual observations and align camera-centric trajectories with a physically meaningful world frame.

To make the reconstructed robot data applicable for dexterous manipulation training, we introduce two components that improve the of grasp-model training and the generalization of manipulation policies. First, we perform (i) differential hand pose optimization by predicting hand-object contact maps to encourage physically plausible interactions, and (ii) interaction-centric grasp modeling to exploit the optimized contact map for valid grasp learning. Second, we leverage DemoGen~\cite{xue2025demogen} to synthesize diverse demonstrations from a single reconstructed video trajectory, enabling one-to-many trajectory generation for generalizable manipulation policy training. We evaluate \mysystem{} on two task categories: object grasping in simulation, and object manipulation in real-world. Comprehensive evaluations and ablations show that the learned grasp model achieves a 70.25\,\% success rate across 20 objects using the Inspire Hand, while the learned manipulation policies with the LEAP Hand achieve an average 62.86\,\% success rate across seven real-world manipulation tasks, including three tasks learned from in-scene videos and four tasks learned from in-the-wild videos.

\noindent \textbf{Contributions.}
\begin{enumerate*}[label=(\roman*), itemjoin=\quad]
\item \mysystem: a framework for learning multi-fingered robotic hand grasping and manipulation directly from RGB human videos alone, without requiring robot demonstrations, wearable devices, or external sensing hardware;
\item An explicit, trainable 3D robot hand-object trajectory reconstruction pipeline from both in-scene and in-the-wild human videos;
\item Using reconstructed robot data, we enable dexterous robot hands to perform diverse object grasping and generalizable manipulation tasks from a single RGB human video per object or task.
\end{enumerate*}

%% file: sec/3-related.tex
\section{Related Work}
\label{sec:related_work}

\subsection{Manipulation Learning from Human Videos}
Large-scale RGB videos of humans interacting with the world are abundant, yet learning dexterous manipulation policies from them remains challenging because human videos do not directly provide robot-executable actions, and hand–object interactions are often ambiguous. Prior works leverages human videos to learn manipulation priors~\cite{shaw_internetvideos, singh2024videos, nairr3m, mavip}, but typically still require robot demonstrations for fine-tuning, limiting generalization from human data alone. Other approaches extract intermediate representations such as object affordances~\cite{shi2025zeromimic, agarwaldexterous, bao2024handsonvlm}, point flows~\cite{ bharadhwaj2024track2act, goyal2022ifor, wangmimicplay, li2025novaflow}, or hand poses~\cite{qin2022dexmv, lum2025crossing}. However, these methods are largely limited to coarse manipulation or gripper-based tasks~\cite{jain2024vid2robot, ye2025video2policy, tang2025mimicfunc, yu2025real2render2real}.

Another line of works demonstrate dexterous manipulation without robot demonstrations, but rely on specialized sensing hardware. AINA~\cite{guzey2025dexterity} uses smart glasses to capture hand–object point clouds, while related methods depend on wearable headsets~\cite{jiang2025dexmimicgen, qiu2025humanoid} or hand motion capture systems~\cite{qin2022dexmv, wang2024dexcap}, introducing additional hardware dependencies. In parallel, world models~\cite{goswami2025world, huang2026pointworld} and video generative models~\cite{li2025novaflow, liang2025video, patel2025robotic} enable zero-shot manipulation by predicting future object trajectories, primarily for gripper-based tasks. LVP~\cite{chen2025large} extends this approach to dexterous hands by retargeting predicted human hand motions, but it does not explicitly model hand-object contacts and can produce infeasible motions. In contrast, our work emphasizes contact-aware grasp modeling and reconstructs reliable, learnable 3D hand-object trajectories from human videos, while remaining complementary to video generation based methods.

\subsection{Human-Object Interaction Reconstruction from Videos} 
Reconstructing hands and objects from images and videos is a long-studied research area. Modern approaches leverage powerful models, such as transformers, to infer low-dimensional hand pose and shape representations~\cite{rong2020frankmocap, pavlakos2024reconstructing}. With strong 3D supervision, it is feasible to learn shared models across multiple object categories using diverse object representations, including meshes~\cite{gkioxari2019mesh}, point clouds~\cite{lin2018learning} and primitives~\cite{deng2020cvxnet}. With these advancements, joint reasoning about hand–object interactions from images and videos has become increasingly feasible~\cite{hasson2019learning, wu2024reconstructing, liu2024easyhoi}, leveraging methods like diffusion models~\cite{ye2023diffusion}, implicit signed distance fields~\cite{park2019deepsdf, fan2024hold} and differentiable pose estimation to optimize the 3D hand-object contact~\cite{cao2021reconstructing, grady2021contactopt}. Recent works have utilized reconstructed hand-object interactions from images and videos to train dexterous robot grasping and manipulation policies~\cite{chen2025web2grasp, singh2024videos, li2025maniptrans}, often by explicitly retargeting human hand motions to robot actions~\cite{qin2023anyteleop, pan2025spider}. Existing methods, however, either require extensive filtering due to reconstruction inaccuracies~\cite{chen2025web2grasp}, incur long reconstruction times~\cite{fan2024hold}, or rely on additional robot demonstrations because reconstructed data alone is insufficient~\cite{singh2024videos}. In contrast, \mysystem learns manipulation policies directly from robot trajectories reconstructed from RGB human videos.

%% file: sec/method.tex
\section{Method}
\label{sec:method}
\begin{figure*}[t]
    \centering
    \includegraphics[width=1.02\linewidth]{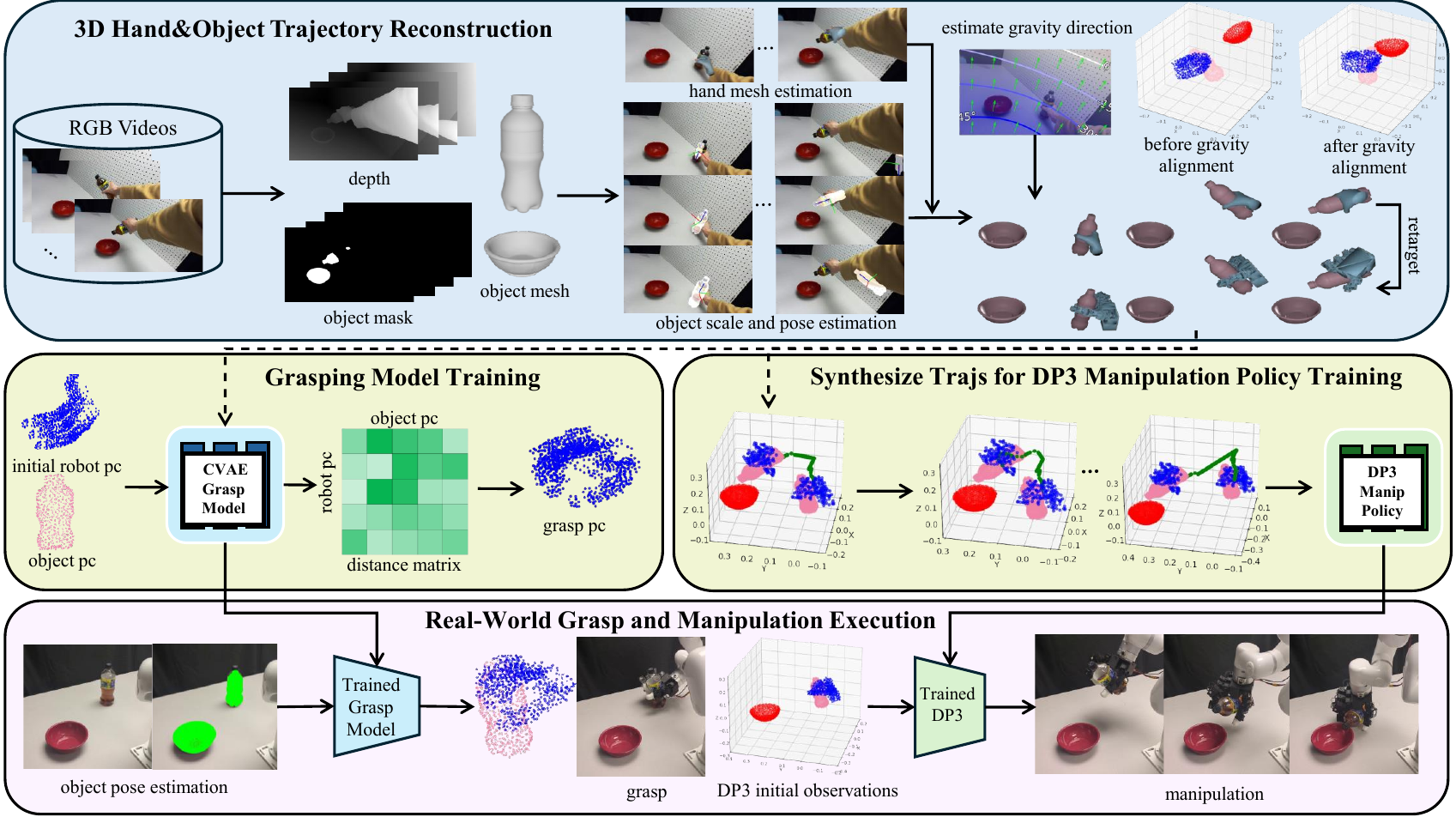}
    \vspace{-10pt}
    \caption{\small \textbf{Overview of the \mysystem framework.} We first reconstruct 3D robot–object interaction trajectories from RGB human videos via recent advances in 3D vision (Sec.~\ref{sec:method_recon}). To utilize the reconstructed data for dexterous grasping and manipulation learning, we perform grasp contact optimization and interaction-centric grasp modeling, and synthesize trajectories for generalizable manipulation (Sec.~\ref{sec:method_train}). Finally, we deploy the trained models for real-world dexterous grasping and manipulation.}
    \label{fig:pipeline}
    \vspace{-10pt}
\end{figure*}

In this section, we present (i) the overall pipeline for reconstructing hand mesh $\mathbf{H}$ and object mesh $\mathbf{O}$ from RGB human videos in Sec~\ref{sec:method_recon}, which yields explicit 3D robot hand trajectories, finger-level manipulations, and the resulting object pose changes; and (ii) the learning of grasping and manipulation policies from these reconstructed hand actions and object point clouds in Sec~\ref{sec:method_train}. 

\noindent \textbf{Assumptions.}
\begin{enumerate*}[label=(\roman*), itemjoin=\quad]
\item The raw dataset consists exclusively of egocentric human videos captured with a static camera, including both in-scene and in-the-wild recordings. Only \textit{one video} is available for each task or object.
\item No pre-scanned object meshes, object size information, depth measurements, or camera intrinsics are assumed to be available. 
\item No robot data or wearable and external sensing devices are used.
\end{enumerate*}

\subsection{3D Hand-Object Trajectory Reconstruction from Video}
\label{sec:method_recon}
Given an RGB video of a human interacting with objects, $\mathcal{V} = \{ I_1, \ldots, I_T \}$, with image frames $I_t \in \mathbb{R}^{H \times W \times 3}$, we leverage existing 3D vision techniques to reconstruct the object mesh $\mathcal{O}$, estimate its 6D pose, and recover the human hand pose, followed by robot hand retargeting. To do so, we first adopt MoGe-2~\cite{wang2025moge} to estimate metric depth maps and camera intrinsics, which define a joint, metric 3D coordinate frame that enables physically consistent hand-object spatial alignment in all subsequent reconstructions.

\textbf{Object Mesh Reconstruction and Pose Estimation.}
Our next step is to reconstruct object mesh $\mathcal{O}$ and its 6D pose from the video.
We reconstruct a complete 3D object mesh directly from RGB observations to enable accurate grasp contact modeling. 
We first identify manipulated objects in the video and obtain object masks using Segment Anything Model~2 (SAM~2)~\cite{ravisam}. We crop the masked regions to produce object-centric images and feed them into MeshyAI for image-to-mesh generation~\cite{meshyai}. While the reconstructed object meshes $\mathcal{O}$ capture the objects’ visual appearance and overall shape, they lack accurate metric scale.Pose estimation methods such as FoundationPose~\cite{wen2024foundationpose} assume access to object meshes with accurate real-world dimensions; violating this assumption leads to erroneous pose estimates.

To address the unknown object scale, we adopt a two-stage scale estimation strategy. First, we query the GPT-4.1 language model to obtain a coarse estimate of the object physical dimension and rescale the mesh accordingly. Second, we further refine the object scale by evaluating multiple candidate scalings (e.g., ranging from 0.5× to 2×) using FoundationPose, leveraging the previously estimated object mask and metric depth information. We select the scaling that minimizes the rendering error, enabling fine-grained scale verification. The rendering error is computed by comparing the rendered object mesh in each frame against the object mask obtained from SAM~2 in the video frames. 

\textbf{Human Hand Mesh Estimation and Robot Hand Retargeting.}
To extract robot actions from video, we first recover human hand motion and then retarget it to the robot hand. We use the hand mesh recovery model HaMeR~\cite{pavlakos2024reconstructing} to reconstruct the hand mesh $\mathcal{H}$ for all video frames. HaMeR represents the hand mesh using a low-dimensional parameterization $\mathbf{h} = (\theta, \beta)$, where $\theta$ denotes the hand pose and $\beta$ denotes the hand shape. 
HaMeR uses a weak-perspective camera model, which is inherently depth-ambiguous and sensitive to focal length errors. To align the reconstructed hand mesh $\mathcal{H}$ within a joint coordinate space with the object, we reuse the metric depth maps predicted by MoGe-2, which are consistent with those used for object pose estimation, to ensure that $\mathcal{H}$ and $\mathcal{O}$ share the same depth reference. We then compute the corrected hand depth $t^{\prime}_z$ by averaging the metric depths at the 2D keypoints predicted by HaMeR. For every frame $I_t$, The reconstructed human hand poses ($\theta_t$ and $\beta_t$) are then retargeted to the robot hand configuration $q_t$, including the wrist pose and finger joint angles, via an optimization that minimizes the error between selected robot link keypoints and their corresponding human hand joints~\cite{qin2023anyteleop}. Given the robot URDF file, the robot mesh $\mathcal{R}$ can be generated for any robot hand configuration $q$.

\textbf{In-the-Wild Videos Calibration.} Trajectories from in-the-wild videos, unlike those from in-scene videos that can be transformed to the robot base using a calibrated transform ${}^{\text{world}}T_{\text{cam}}$, lack this transformation. Their actions are expressed in the camera frame and are misaligned with the world frame due to unknown camera orientation and pose.
As illustrated in Figure~\ref{fig:pipeline}, before gravity alignment the reconstructed objects (e.g., the bowl and bottle) do not lie on the horizontal plane due to a tilted camera. Policies trained on such data would therefore learn camera-frame specific actions that do not transfer to the physical world frame. To address this issue, we apply GeoCalib~\cite{veicht2024geocalib}, a single-image camera calibration method that leverages universal visual cues of 3D geometry to estimate camera orientation. From the first video frame, GeoCalib infers the gravity direction in the camera frame and the corresponding rotation ${}^{\text{grav}}R_{\text{cam}} \in \mathrm{SO}(3)$ that aligns gravity with the negative \(z\)-axis, $\mathbf{g}_{\text{cam}} = [0, 0, -1]^\top$. We apply this rotation to all reconstructed meshes $\mathcal{R}$, $\mathcal{O}$, and robot configurations $q$, producing gravity-aligned trajectories and point clouds suitable for policy training. While this does not recover the full camera-to-world transformation ${}^{\text{world}}T_{\text{cam}}$, the estimated ${}^{\text{grav}}R_{\text{cam}}$ is sufficient to align trajectories from egocentric, in-the-wild videos to a shared reference plane with in-scene videos (e.g., the robot-mounted table), yielding trajectories that closely match those reconstructed from in-scene recordings.

\subsection{Dexterous Grasp and Manipulation Learning}
\label{sec:method_train}
Here, we discuss how to exploit the reconstructed trajectories to teach robot dexterous manipulation. To handle visual differences in hand embodiment and background environments between robot and human RGB videos, we adopt point-cloud–based policies by sampling points from meshes' surfaces. In practice, although the reconstructed trajectories capture the correct motions, the resulting robot–object interactions are not always physically feasible due to reconstruction errors in object geometry and inaccuracies in hand mesh scale, shape, or pose, which can lead to interpenetration or invalid contact (see Figure~\ref{fig:grasp_info}(b) for examples). During execution, such errors may cause the robot to miss the object or fail to grasp it securely. Moreover, although a single human video may include multiple grasp instances, it yields only one manipulation trajectory, which is insufficient for learning robust manipulation policies.

To improve the physical validity of hand-object interactions and increase trajectory diversity from a single video, we perform differentiable hand–object contact optimization~\cite{grady2021contactopt} and adopt DemoGen~\cite{xue2025demogen} to synthesize spatially randomized manipulation demonstrations from a single reconstructed human trajectory.
Following DemoGen’s skill–motion decomposition, we split the reconstructed trajectory into two stages, grasping and manipulation, each learned by a dedicated policy. The grasp stage spans from the time the hand approaches sufficiently close to the object at time $t_1$ to when a stable grasp is achieved at time $t_2$, before any object manipulation begins. The manipulation stage covers $[t_2, T]$, where the hand operates on the grasped object. The grasping policy is responsible for learning correct hand–object interactions, whereas the manipulation policy controls the object’s relative translation and rotation along with fine-grained finger motions.

\textbf{Contact Optimization and Interaction-Centric Grasp Modeling.} 
To address grasp reconstruction failures, during the grasping stage $[t_1, t_2]$, we refine the grasps using pre-trained ContactOpt~\cite{grady2021contactopt}.The hand parameters $\mathbf{h}$ determine the geometry of the reconstructed hand mesh $\mathcal{H}$ relative to the object mesh $\mathcal{O}$. We therefore compute hand pose-dependent contact maps $C_\mathcal{H}(\mathbf{h})$ and $C_\mathcal{O}(\mathbf{h})$, which assign smooth, distance-based contact values to each vertex based on its proximity to the other mesh. For example, the contact value at an object vertex $v_\mathcal{O}^i$ can be expressed as
\[
C_\mathcal{O}(v_\mathcal{O}^i; \mathbf{h}) = \max\Big(0, 1 - \frac{\min_j \| v_\mathcal{H}^j(\mathbf{h}) - v_\mathcal{O}^i \|}{c_\text{rad}} \Big),
\]
where $v_\mathcal{H}^j(\mathbf{h})$ are the hand vertices at pose $\mathbf{h}$ and $c_\text{rad}$ is a radius parameter controlling the falloff. Similarly, $C_\mathcal{H}(\mathbf{h})$ is computed for the hand vertices. These maps are differentiable and provide gradients for optimizing the hand pose.
ContactOpt then predicts desirable contact regions on the hand mesh $\hat{C}_\mathcal{H}$ and object mesh $\hat{C}_\mathcal{O}$, and adjusts the hand pose parameters $\mathbf{h}$ to align the current contacts with these targets using the differentiable objective (for the full formulation, please refer to~\cite{grady2021contactopt}):
\[
E(\mathbf{h}) = \lvert C_\mathcal{O}(\mathbf{h}) - \hat{C}_\mathcal{O} \rvert + \lvert C_\mathcal{H}(\mathbf{h}) - \hat{C}_\mathcal{H} \rvert.
\]
The optimized human hand poses are then retargeted to robot hand configurations ${q^{\text{opt}}_t}, t \in [t_1, t_2]$, which serve as grasp demonstrations for training the grasping model.

Given the robot mesh $\mathcal{R}$ from $q^{\text{opt}}_t$ and the object mesh $\mathcal{O}$, we then employ the DRO model~\cite{wei2025dro} to capture their interaction for robust grasp modeling. DRO predicts dense point-to-point distances between the robot hand point cloud $\mathbf{P}^{\mathcal{R}} \in \mathbb{R}^{N_{\mathcal{R}} \times 3}$ and the object point cloud $\mathbf{P}^{\mathcal{O}} \in \mathbb{R}^{N_{\mathcal{O}} \times 3}$, effectively capturing the interaction patterns and spatial relationships provided by ContactOpt.
Specifically, DRO takes randomly initialized robot hand point clouds $\mathbf{P}_{\text{init}}^{\mathcal{R}}$ and zero-centered object point clouds $\mathbf{P}^{\mathcal{O}}$ sampled from the meshes as input, and predicts the distance matrix $\mathcal{D}(\mathcal{R}, \mathcal{O})^{\text{Pred}} \in \mathbb{R}^{N_{\mathcal{R}} \times N_{\mathcal{O}}}$, as illustrated in Figure~\ref{fig:pipeline}.
The training loss is the difference between the predicted and ground-truth distance matrices with $\mathcal{L_{\text{L1}}}\left( \mathcal{D(R,O)}^{\text{Pred}}, \mathcal{D(R,O)}^{\text{GT}} \right)$. Using $\mathcal{D(R,O)}^{\text{Pred}}$ and the object point clouds $\mathbf{P}^{\mathcal{O}}$, we can position the robot point cloud in the target grasp pose $\mathbf{P}_{\text{grasp}}^{\mathcal{R}}$ with a multilateration method~\cite{norrdine2012algebraic} and compute the grasp configuration $q^{\text{grasp}}$ through optimization, expressed relative to the object.

\textbf{Manipulation Demonstration Synthesis and Training.}
Generalizable manipulation policies typically require a large number of demonstrations, whereas our goal is to learn from a single reconstructed human video to avoid repetitive human video collection and reconstruction. To this end, we adopt DemoGen~\cite{xue2025demogen} to synthesize additional manipulation demonstrations via objects spatial randomization. DemoGen enforces spatial equivariance by applying consistent SE(3) transformations to both the object point clouds and the associated robot trajectories. Since 3D point clouds and robot proprioceptive states can be transformed directly, this enables the synthesis of diverse trajectories while preserving hand–object contact and fine-grained finger motions.
For manipulation policy learning, we adopt the 3D diffusion-based policy (DP3)~\cite{Ze2024DP3}, which takes as input the robot hand point clouds and proprioceptive state at the grasp pose, $\mathbf{P}_{\text{grasp}}^{\mathcal{R}}$ and $q^{\text{grasp}}$, along with the object point clouds $\mathbf{P}^{\mathcal{O}}$ as the initial observation. The policy outputs actions as predicted changes in the robot configuration, $\Delta q$, and is executed in a closed-loop manner using updated observations.

%% file: sec/experiment.tex
\section{Experiments}
\begin{figure}[t]
    \centering
    \includegraphics[width=1.0\linewidth]{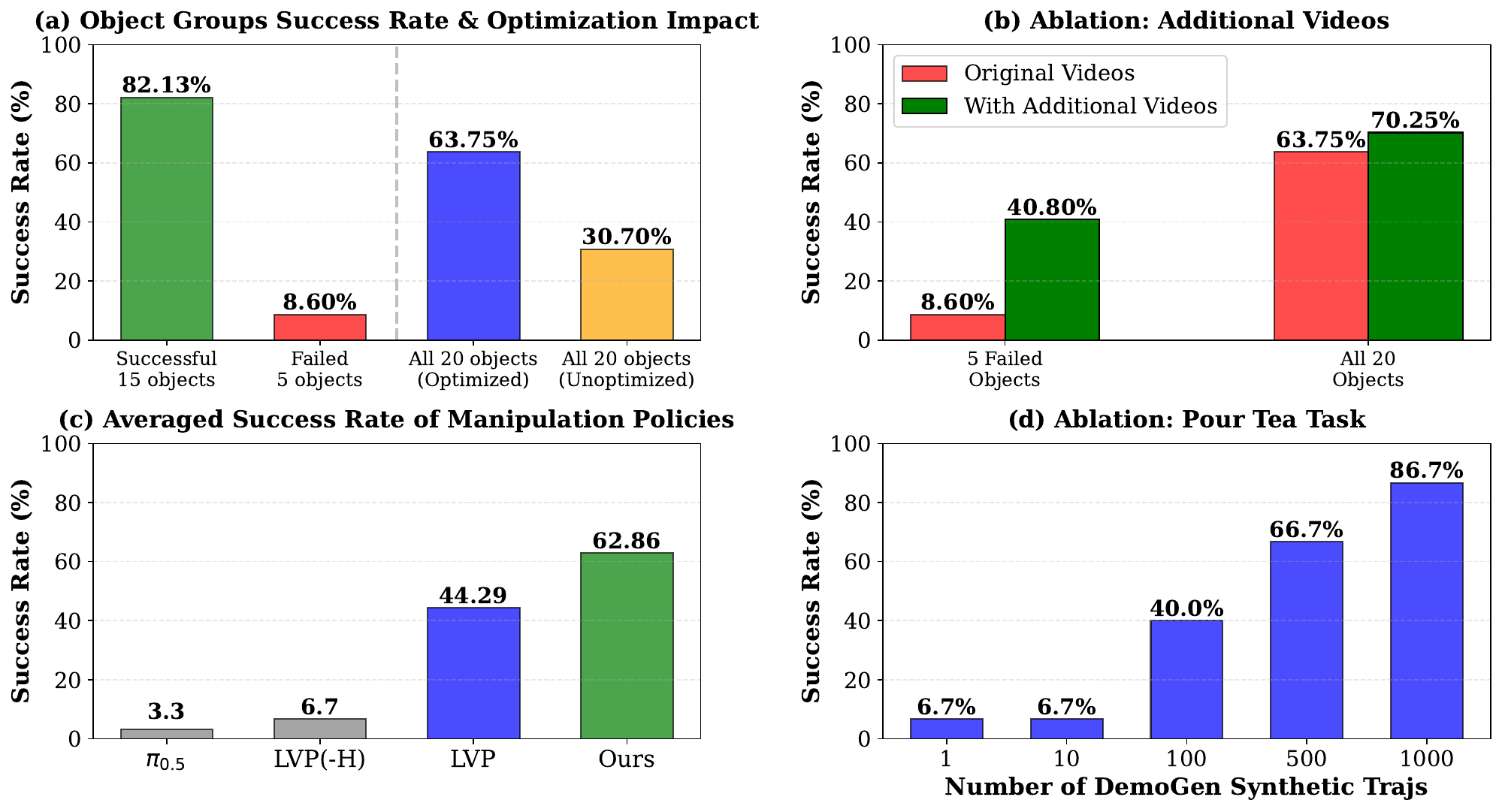}
    \vspace{-8pt}
    \caption{\small \textbf{Quantitative Results on Grasping and Manipulation.} \textbf{Grasping:} (a) Success rates across object groups, with comparison between models trained with and without grasp optimization; (b) Ablation study on incorporating additional videos for previously failed objects. \textbf{Manipulation:} (c) Performance comparison between our method and baselines across seven manipulation tasks; (d) Ablation study on the number of DemoGen-synthesized trajectories.}
    \label{fig:overall_result}
    \vspace{-8pt}
\end{figure}

We conduct two sets of experiments with our framework: grasping of objects and manipulating different objects. We present the main results in Fig.~\ref{fig:overall_result}. Video rollouts for the results are best viewed in our website.

\subsection{Grasping Experiments}
\begin{figure}[t]
    \centering
    \includegraphics[width=0.87\linewidth]{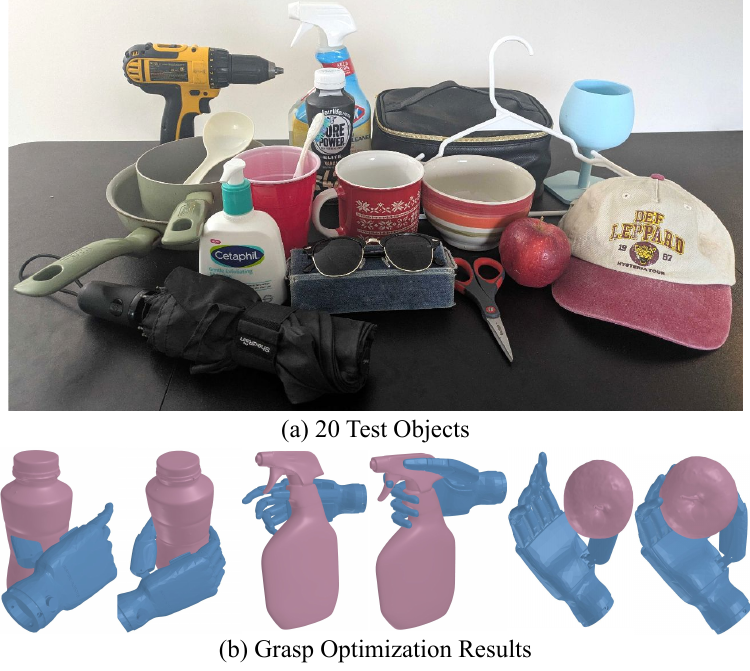}
    \vspace{-5pt}
    \caption{\small
    (a) Visualization of the 20 objects used for video collection and grasp model training.
    (b) Grasp optimization results on \textit{Bottle}, \textit{Spray Bottle}, and \textit{Apple}. Left: unoptimized reconstructed grasps; right: grasps optimized using ContactOpt~\cite{grady2021contactopt}.}
    \label{fig:grasp_info}
    \vspace{-10pt}
\end{figure}

\textbf{Setup.} We collect 20 videos of humans grasping objects, with one video per object; the object categories are shown in Fig.~\ref{fig:grasp_info}~(a). Each 30~fps video lasts 3 to 6 seconds. We train our models across all objects and evaluate grasp success in the IsaacGym simulator using an 18-Dof Inspire robot hand, as its size closely matches that of a human hand.
We initialize the simulator with object meshes and poses reconstructed from the recorded videos. Each grasp $q^{\text{grasp}} \in \mathbb{R}^{18}$ undergoes a 300-step disturbance phase in IsaacGym, with forces applied sequentially from six directions: $\pm x$, $\pm y$, and $\pm z$. The applied force magnitude is set to 0.5 times the object's mass. We consider a grasp successful if the object’s displacement under applied force disturbances stays within 3\,cm of its initial position. Through these experiments, we aim to answer the following questions:
\begin{itemize}
\item \textbf{Q1}: How well does the grasping model generalize across object categories when trained on grasps reconstructed from human RGB videos?
\item \textbf{Q2}: For objects that are initially challenging to grasp, does incorporating additional videos captured from diverse viewpoints improve performance?
\end{itemize}

\begin{figure*}[t]
    \centering
    \includegraphics[width=1.0\linewidth]{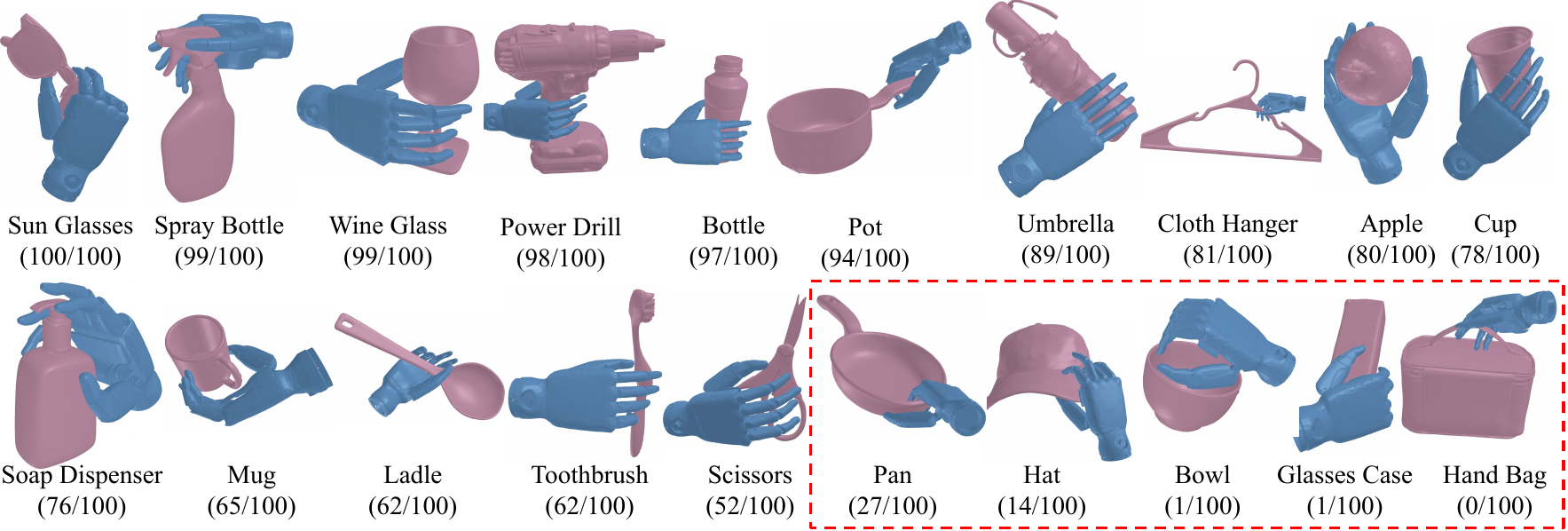}
    \vspace{-10pt}
    \caption{\small \textbf{Predicted grasps and success rates in IsaacGym.} The DRO grasping model is trained on 20 object categories. Each object is evaluated over 100 trials and sorted by descending success rate; red dotted box denotes failed grasps.}
    \label{fig:grasp_vis}
    \vspace{-10pt}
\end{figure*}

\textbf{A1:} Our grasping model, trained on video-reconstructed grasp data, achieves an average success rate of 82.13\% over 15 successfully grasped objects and 63.75\% across all objects, including five failure cases, as shown in Fig.~\ref{fig:grasp_vis}.
The predicted grasps closely mimic the human grasps observed in the input videos; for example, on \textit{Soap Dispenser} and \textit{Spray Bottle}, the middle finger is placed on top of the dispenser and the trigger, respectively. We further ablate the effect of grasp optimization using ContactOpt~\cite{grady2021contactopt} by retraining the DRO grasping model on all objects with unoptimized grasps, which yields only a 30.7\% average success rate across 20 objects, compared to 63.75\% when using optimized grasps, as shown in Fig.~\ref{fig:overall_result}(a). As shown in Fig.~\ref{fig:grasp_info}(b), unoptimized grasps often penetrate object surfaces or fail to establish contact due to reconstruction errors, whereas optimized grasps achieve more accurate hand–object contact, which is reflected in the higher grasp success rate.

\textbf{A2:} However, five objects (highlighted by the red dotted box in Fig.~\ref{fig:grasp_vis}) exhibit lower success rates. These failures primarily stem from object pose estimation errors, for example in the \textit{Pan} case where heavy occlusion of the handle by the human hand leads to inaccurate pose estimation, and from a lack of force awareness, in which grasps for the \textit{Glasses Case} and \textit{Hand Bag} are kinematically plausible but fail to achieve force closure, resulting in instability. 
To mitigate these issues, we examine whether incorporating grasping videos with diverse viewpoints and grasp styles improves performance. The underlying intuition is that reconstruction may fail in some videos due to occlusion or suboptimal grasps, but succeed in others with reduced occlusion or more stable force-closure configurations. Accordingly, for each failed object, we collect two additional human videos and augment the dataset, resulting in 30 videos (20 + 5$\times$2) for retraining. As shown in Tab.~\ref{tab:grasp_multi_video}, training with the augmented dataset improves the success rate on the five previously failed objects from 8.6\% to 40.8\%, and increases the overall success rate from 63.75\% to 70.25\%. Although the \textit{Hand Bag} does not improve due to its grasp being sensitive to disturbances, the performance gains on the remaining objects indicate that increased video diversity alone can substantially enhance model robustness and generalization.

\begin{table}[t]
\centering
\caption{\small \textbf{Quantitative ablation of grasp performance with multiple videos.}
Success rates on the five initially failed objects in Fig.~\ref{fig:grasp_vis}, comparing models trained with the original 20 videos and with additional videos for the failed objects.}
\small
\scalebox{0.95}{
\begin{tabular}{lccccc}
\toprule
& Pan& Hat& Bowl& Glasses Case& Hand bag \\
\midrule
Single Video&27& 14& 1& 1& 0 \\
Multi Videos& 36&52&48 &68& 0\\

\bottomrule
\end{tabular}
}
\label{tab:grasp_multi_video}
\vspace{-5pt}
\end{table}

\subsection{Manipulation Experiments}
\textbf{Setup.} We evaluate seven manipulation tasks on a four-finger LEAP Hand mounted on a 7-DoF xArm, using one human demonstration video per task. Three tasks use in-the-scene videos: \textit{Pour Tea} (pick up a bottle and tilt to pour into a bowl), \textit{Close Drawer} (push a drawer fully into a shelf), and \textit{Pick\&Place Can} (pick up a can and place it into a box). The remaining four tasks use in-the-wild videos with gravity-based calibration: \textit{Pour Tea} without camera–robot calibration, \textit{Hang Hat} (hang a hat held in the hand onto a rack), \textit{Move Jenga Box} (pick a Jenga box and place it onto a shelf), and \textit{Screw Bulb} (screw a bulb into a socket until it's on). For \textit{Pick\&Place Can} task, the DP3 action $\Delta q_t$ and state $q_t$ spaces are 22-DoF, comprising a 6-DoF delta wrist pose and 16-DoF delta finger joints to enable object release. For all other tasks, we use a 6-DoF delta wrist action space. We train a single DRO model across all grasp objects, including the \textit{Bottle}, \textit{Can}, \textit{Drawer}, \textit{Hat}, \textit{Jenga Box}, and \textit{Bulb}, and train a separate DP3 manipulation policy for each task. The second object in each task (e.g., \textit{Bowl}, \textit{Box}, \textit{Rack}) is referred to as the target object.

\begin{table}[t]
    \centering
    \caption{\small \textbf{Quantitative comparison of our method and baselines across different video sources and tasks over 10 trials with randomized target object locations.} Our DP3 model is trained on 1000 trajectories synthesized by DemoGen from a single reconstructed source trajectory.}
    \label{tab:manip_comparison}
    \begin{tabular}{cc|ccccc}
        \toprule
        \textbf{Video Type} & \textbf{Task} & $\pi_{0.5}$ & LVP(-H) & LVP & Ours \\
        \midrule
        \multirow{3}{*}{\makecell{In-scene\\ Videos}}
            & Pour Tea & 0/10 & 1/10 & 7/10 & \textbf{8/10} \\
            & Close Drawer & 1/10 & 2/10 & 6/10 & \textbf{9/10} \\
            & Pick\&Place Can & - & 0/10 & 4/10 & \textbf{5/10}\\
        \midrule
        \multirow{4}{*}{\makecell{In-the-wild\\ Videos}}
            & Pour Tea & 0/10 & 1/10 & \textbf{7/10} & \textbf{7/10}\\
            & Hang Hat & 0/10 & - & 4/10 & \textbf{6/10} \\
            & Screw Bulb & 1/10 & 0/10 & 1/10 & \textbf{4/10} \\
            & Move Jenga Box & 0/10 & 0/10 & 2/10 & \textbf{5/10}\\
        \bottomrule
    \end{tabular}
\end{table}

During evaluation, we estimate object poses using FoundationPose~\cite{wen2024foundationpose} with the reconstructed object mesh $\mathcal{O}$ under a calibrated ZED camera. The trained DRO model then predicts a grasp pose from the object point cloud, and the LEAP Hand executes the grasp $q^{\text{grasp}} \in \mathbb{R}^{22}$ in free space. The initial observation for the subsequent manipulation stage comprises the point clouds of the grasped and target object $\mathbf{P}^{\mathcal{O}}$, the robot hand point cloud at the predicted grasp pose $\mathbf{P}_{\text{grasp}}^{\mathcal{R}}$, and the corresponding proprioceptive state $q^{\text{grasp}}$. DP3 is then rolled out in a closed-loop manner based on the current observation. At each timestep, the predicted action $\Delta q_t$ updates the robot configuration and the associated hand point clouds. During execution, the grasped object point cloud $\mathbf{P}^{\mathcal{O}}$ is often occluded by the large LEAP Hand, making closed-loop DP3 control difficult. We therefore assume that the hand-object relative pose established at grasp remains fixed during rollout, and update the object point cloud using the hand-to-object transformation computed from $q^{\text{grasp}}$. This approximation suffices for most tasks, while object release in \textit{Pick\&Place Can} is handled implicitly during execution rather than explicitly modeled in the point-cloud update.

Since our policies are trained solely from RGB human videos without any robot demonstrations, there are few directly comparable baselines. We therefore consider two representative alternatives: (1) \textit{$\pi_{0.5}$}, a generalizable vision–language–action model~\cite{intelligence2025pi} used as a baseline for zero-shot manipulation; and (2) Large Video Planner (LVP), a retargeting-based approach that maps predicted human hand motions from its video generator to robot actions~\cite{chen2025large}.
We fine-tune $\pi_{0.5}$-DROID using 200 robot demonstrations in our scenes and LEAP Hand on tasks disjoint from the seven tasks. Since $\pi_{0.5}$-DROID is trained with parallel-jaw grippers, we initialize each rollout with the object already grasped by the LEAP Hand and evaluate only the subsequent manipulation phase.
For LVP, we evaluate two variants: LVP, conditioned on an initial observation containing a human hand to predict subsequent hand grasping and manipulation motion, and LVP(-H), conditioned on an initial observation containing only the objects, and is used to evaluate the model’s ability to infer manipulation purely from object observations without human hand cues.
Through the experiments, we aim to answer the following questions:
\begin{itemize}
    \item \textbf{Q3}: How well does the manipulation policy perform when trained from a single in-scene and in-the-wild RGB human video?
    \item \textbf{Q4}: How do in-the-wild calibration and the number of synthesized demonstrations affect policy performance?
\end{itemize}

\begin{figure*}[t]
    \centering
    \includegraphics[width=0.99\linewidth]{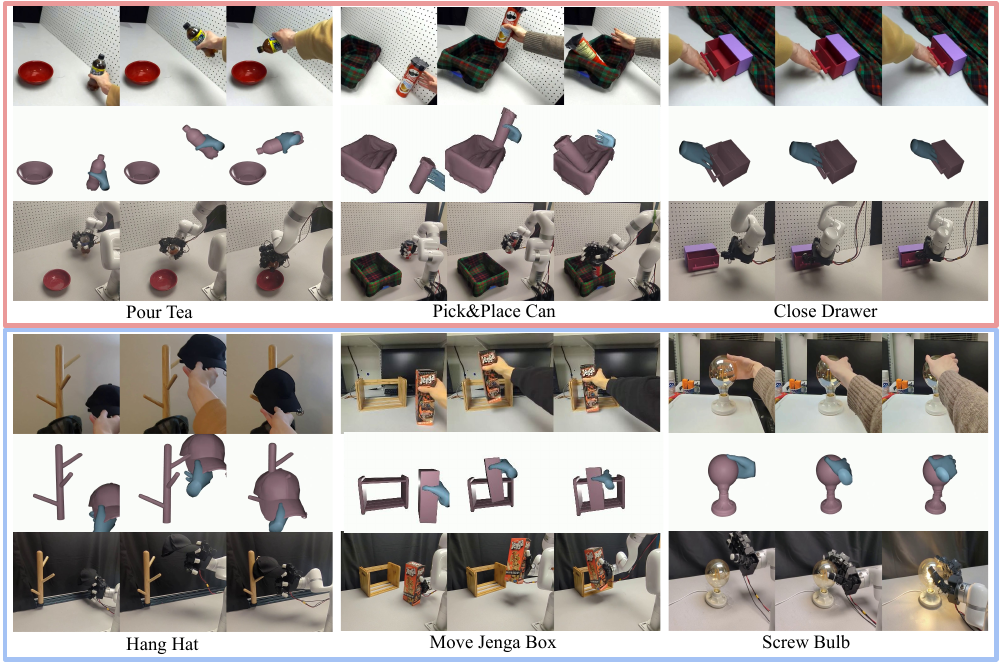}
    \vspace{-5pt}
\caption{\small \textbf{Visualization of \mysystem Execution Using In-Scene (top) and In-the-Wild (bottom) Video Data Sources.} For each task, given RGB human videos (row 1), we reconstruct 3D trajectories of the human hand and objects (row 2). Trained with these trajectories, we executed on a real-world LEAP Hand (row 3).}
    \label{fig:manip_vis}
    \vspace{-10pt}
\end{figure*}

\textbf{A3:} \mysystem captures hand-object motion from human videos, providing spatiotemporal supervision for manipulation policy learning. The reconstructed trajectories closely align with the corresponding RGB human videos, as shown in Fig.~\ref{fig:manip_vis}. Leveraging both these reconstructed trajectories and additional synthesized trajectories generated by DemoGen, the trained DP3 policy achieves the highest overall success rate of 62.86\% across all tasks (Tab.~\ref{tab:manip_comparison}). The synthesized trajectories enable our policy to generalize towards different target object locations(videos available on our website). While, DemoGen based trajectory augmentation is less helpful for tasks like  \textit{Close Drawer} and \textit{Screw Bulb}, where objects (\textit{Bulb} and \textit{Socket}) are tightly coupled, the variations in object locations are still helpful for generalization of the trained policy.

Besides, our model outperforms the LVP baselines, which fail to consistently generate feasible grasps for the \textit{Jenga Box} and exhibit motion failures in tasks such as \textit{Move Jenga Box} (object not placed on the shelf) and \textit{Screw Bulb} (missing required rotational motion). Moreover, when the initial observation lacks a human right hand, a common scenario in robot manipulation, LVP(-H) frequently generates erroneous future frames even when explicitly instructed to use the right hand. These failures include hallucinating a robot gripper instead of a human hand, failing to execute the intended task (e.g., remaining static instead of push the drawer), attempting infeasible grasps, confusing left and right hands, or producing undetectable hands. In contrast, our approach reconstructs explicit 3D geometry, grounding supervision in physically meaningful meshes rather than pixel-level video hallucinations. Together with synthesized trajectories, our results demonstrate that reconstruction-based trajectories provide reliable and generalizable supervision for dexterous manipulation policy learning. Finally, $\pi_{0.5}$ achieves limited success because it is trained primarily on parallel-gripper datasets, and the small scale of our LEAP Hand fine-tuning demonstrations is insufficient to meaningfully improve its dexterous manipulation performance.

\textbf{A4:} The \textit{Pour Tea} task is evaluated on both in-scene and in-the-wild video categories using ${}^{\text{world}}T_{\text{cam}}$ and ${}^{\text{gra}}R_{\text{cam}}$, respectively. Table~\ref{tab:manip_comparison} shows no significant performance difference between the two video sources; however, without applying ${}^{\text{gra}}R_{\text{cam}}$ for the in-the-wild \textit{Pour Tea} task, the success rate drops to 0\%, highlighting the importance of gravity-alignment calibration. We further ablate the effect of DemoGen on the in-scene \textit{Pour Tea} task by adding $x$-$y$ plane perturbations within $[-0.2, 0.2]$\,m to the target object \textit{Bowl} position reconstructed from the source video, and synthesizing between 1 and 1000 trajectories. During evaluation, the \textit{Bowl} is placed at 15 different locations within this perturbed region. As the number of synthesized trajectories increases, the success rate improves steadily from 1/15 to 13/15 (Fig.~\ref{fig:overall_result}(d)), demonstrating that DemoGen enables DP3 to learn actions that generalize across target object locations.